\documentclass{article}
\usepackage{iclr2027_conference,times}
\usepackage[T1]{fontenc}
\usepackage{microtype}
\usepackage{amsmath,amssymb,booktabs,graphicx,array,longtable}
\usepackage{xcolor}
\usepackage{hyperref}
\usepackage{url}
\hypersetup{
  hidelinks,
  pdftitle={SchemaMem: Schema-Indexed Recurrent Memory for Delayed State Retrieval},
  pdfauthor={Sungwoo Goo, Hwi-yeol Yun, Sangkeun Jung}
}
\graphicspath{{figures/}}
\newcommand{\sm}{SchemaMem}
\newcommand{\softmax}{\operatorname{softmax}}
\newcommand{\norm}{\operatorname{RMSNorm}}
\title{SchemaMem: Schema-Indexed Recurrent\\Memory for Delayed State Retrieval}
\author{%
  Sungwoo Goo$^{1}$ \qquad Hwi-yeol Yun$^{1}$ \qquad Sangkeun Jung$^{2}$\\
  \small $^{1}$College of Pharmacy, Chungnam National University\\
  \small $^{2}$Department of Computer Science \& Engineering, Chungnam National University\\
  \small Daejeon, Republic of Korea\\
  \small \texttt{swgoo91@gmail.com, hyyun@cnu.ac.kr, hugmanskj@gmail.com}%
}

\iclrfinalcopy

\begin{document}
\maketitle
\lhead{Preprint}
\begin{abstract}
Attention provides direct access to past representations, but retaining an ever-growing history is costly.
Recurrent models bound persistent state, yet must preserve selected information while processing subsequent inputs.
We introduce \sm{}, an attention-based recurrent memory architecture combining chunk-local attention with a persistent, schema-indexed phase state.
Learned schema embeddings provide a shared representational reference for reading and
writing.
Reads use the current state, whereas writes use the layer input and static schema embeddings, excluding direct feedback from that layer's own state.
Chunk-boundary commits aggregate bounded phase increments through forward computation.
The same parameters also support full-history attention training before and during recurrent training.
We studied selective updates, preservation, and delayed retrieval in a controlled address--value task, comparing three-layer models with approximately matched parameter counts and persistent-state dimensions. Across nine address/value settings and three training seeds, \sm{} has higher mean written-value retention at four times
the maximum training delay than both baselines, which are trained toward a higher in-range accuracy target.
Updated-value recovery favors \sm{} in all nine settings against Mamba-3 and seven against Gated DeltaNet.
Defaults consistently favor Gated DeltaNet over \sm{} at that delay, and \sm{} requires substantially more optimization steps.
These results identify a promising retention--optimization trade-off in schema-indexed recurrence.

\end{abstract}
\section{Introduction}
\label{sec:intro}
Maintaining a useful state is different from retaining an input history.
A system tracking an evolving collection of facts must incorporate changes, leave unrelated facts intact, and recover the current value after intervening activity.
Full-history attention can consult the original observations \citep{vaswani2017attention}; a bounded-state recurrent model instead has to encode their consequences in a representation that survives subsequent computation.
Good retrieval immediately after a write does not by itself establish this latter ability.

Modern state-space and recurrent linear-attention models offer powerful alternatives to growing attention histories \citep{gu2023mamba,dao2024ssm,lahoti2026mamba3,yang2024gated}.
However, their performance on a learned sequence distribution does not determine how well selected updates survive longer streams of intervening inputs.
We investigate that distinction using a small, controlled task with exact answers.
The task is intentionally simpler than natural language: it isolates preservation and revision of address--value associations, without conflating them with language comprehension or generation.

Our architectural proposal is \sm{}.
Each memory layer combines a learned schema bank $E$ with an episode-specific phase state $S$.
Attention reads from the combined representation, while an independent writer uses the incoming hidden representations and the static bank to form additive phase increments.
These increments are committed at chunk boundaries.
The schema bank is not an empty set of storage cells: it contributes a learned representation even when $S=0$.
Conversely, $S$ changes the bank without changing model parameters during inference.

The design separates two problems.
Chunk-local attention processes recent representations directly; cross-chunk communication must use the persistent state.
Excluding same-layer state from the writer also separates the explicit state carry from its readout, while still permitting lower-layer memory to influence higher-layer writes.
The resulting implementation uses standard attention, linear projections, elementwise functions, and matrix-product aggregation.
It does not require a model-specific recurrent scan kernel.

We evaluate this design against Mamba-3 and Gated DeltaNet at roughly 55,000 parameters and 17,000 persistent-state elements. 
All three models must learn the same database semantics.
We then increase the delay between updates and final queries without further training.
The central observation is a \emph{delay-dependent reversal}: baselines can be more accurate within the training range, yet retain fewer written values at substantially longer delays.
This reversal is clearest in the 8- and 16-address settings, is not uniform at higher loads, and comes with a large training-cost disadvantage for \sm{}.

Our contributions are: 
(i) an attention-compatible recurrent memory with a learned schema reference, state-independent same-layer writing, and additive
phase commits;
(ii) a controlled evaluation separating unchanged written values, new values, and never-written defaults;
and (iii) evidence that high in-range accuracy need not imply comparable delayed state retention, together with the optimization and capacity limitations of the proposed alternative.
\section{Related Work}
\label{sec:related}
\paragraph{Attention with persistent and recurrent memory.}
Neural Turing Machines use differentiable addressing to read and write an external memory \citep{graves2014ntm}.
Learned persistent memory vectors also provide useful input-independent representations in attention networks \citep{sukhbaatar2019persistent}.
Transformer-XL introduces segment recurrence, and Compressive Transformers retain compressed past representations
\citep{dai2019transformerxl,rae2019compressive}.
Recurrent Memory Transformers pass memory tokens between segments \citep{bulatov2022rmt}.
Block-Recurrent Transformers are a particularly close architectural precedent:
they use attention over blocks and a large recurrent state with gating \citep{hutchins2022block}.
Our distinction is the combination of a learned schema reference with episode-specific phase residuals, a static-schema writer, and additive boundary commits.

\paragraph{State-space and fast-weight sequence models.}
Linear attention admits a recurrent formulation \citep{katharopoulos2020linear}.
Mamba makes state-space dynamics input-selective, and Mamba-2 develops a structured state-space duality with attention \citep{gu2023mamba,dao2024ssm}. 
Mamba-3 further develops its discretization, complex-valued state update, and inference-oriented design \citep{lahoti2026mamba3}.
DeltaNet and Gated DeltaNet use delta-rule memory updates, with the latter combining targeted modification and gating \citep{yang2024delta,yang2024gated}. These are expressive learned memories, not merely fixed exponential filters.

\paragraph{Memory persistence as a separate evaluation axis.}
MemMamba analyzes memory decay in selective state-space models and augments them with summarization and attention \citep{wang2025memmamba}.
This motivates testing persistence after intervening computation rather than immediate retrieval alone.
Our task separates different kinds of state changes and reports exact-value as well as bit accuracy.
\section{Schema-Indexed Recurrent Memory}
\label{sec:method}
Each layer has a learned schema bank $E\in\mathbb{R}^{M\times d}$ and an episode-specific phase state $S_c$ of the same shape. 
$S_c$ starts at zero, is fixed within chunk $c$, and is not a model parameter.
Banks are independent across layers, but each layer shares its bank projections between reading and writing.
A schema slot is not assigned to a database address by the implementation.

\begin{figure}[t]
\centering
\includegraphics[width=\linewidth]{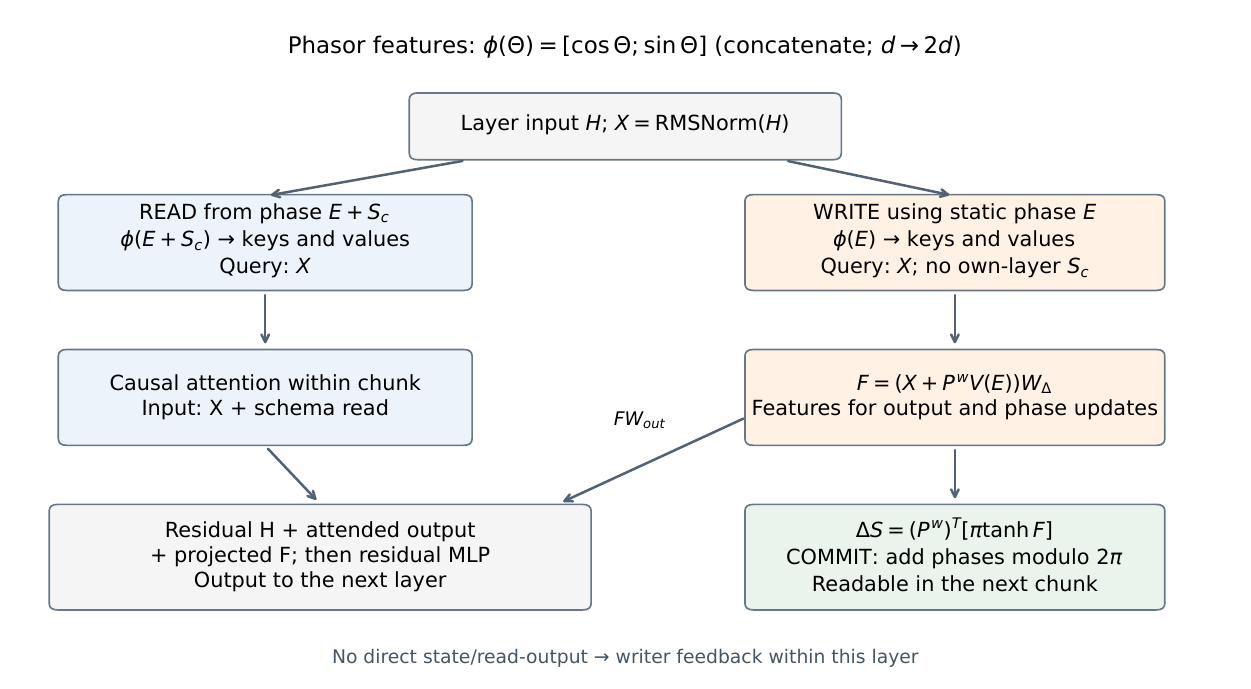}
\caption{One \sm{} layer. The phasor feature map
$\phi(\Theta)=[\cos\Theta;\sin\Theta]$ concatenates elementwise cosine and
sine along the feature dimension ($d\to2d$). Learned projections map these
features to keys and values. Reading uses phase $E+S_c$; writing uses the
pre-read layer input and static phase $E$ only. The writer excludes direct
same-layer state feedback to mitigate gradient instability observed in
preliminary experiments. Local attention cannot access
previous chunks' keys or values. Commits make updates readable in the next
chunk, and lower-layer memory can influence higher-layer writes through the
hidden representation.}
\label{fig:architecture}
\end{figure}

\paragraph{Schema reads and independent writes.}
For a chunk of $C_c$ positions, let $H\in\mathbb{R}^{C_c\times d}$ be the
layer input and $X=\norm(H)$, omitting the batch axis.
The phasor map $\phi(\Theta)=[\cos\Theta;\sin\Theta]$ concatenates
elementwise cosine and sine along the feature dimension.
Learned projections $W_K,W_V\in\mathbb{R}^{2d\times d}$ give
$K(\Theta)=\phi(\Theta)W_K$ and $V(\Theta)=\phi(\Theta)W_V$.
These projections are shared by the read and write branches, while their
query projections $W_{\rm r},W_{\rm w}\in\mathbb{R}^{d\times d}$ are distinct.
Suppressing layer indices, the core operations are
\begin{align}
 P^{\rm r}&=\softmax_M\!\left((XW_{\rm r})K(E+S_c)^\top/\sqrt d\right),
 & R&=P^{\rm r}V(E+S_c), \label{eq:core-read}\\
 P^{\rm w}&=\softmax_M\!\left((XW_{\rm w})K(E)^\top/\sqrt d\right),
 & F&=(X+P^{\rm w}V(E))W_\Delta, \label{eq:core-write}\\
 \Delta S_c&=(P^{\rm w})^\top[\pi\tanh(F)],
 & S_{c+1}&=(S_c+\Delta S_c)\bmod 2\pi. \label{eq:core-commit}
\end{align}
The softmax is over the $M$ schema slots for each input position, so both
addressing matrices $P^{\rm r},P^{\rm w}\in\mathbb{R}^{C_c\times M}$ are dense.
Reading uses $E+S_c$: the episode-specific state changes both the keys used
to locate information and the values returned by the weighted read.
Even at $S_c=0$, $R$ supplies an input-dependent mixture of learned static
schema values rather than an empty memory read.

Writing instead queries the static bank $E$ with the pre-read input $X$.
$P^{\rm w}$ determines where to distribute each position's update, while
$F\in\mathbb{R}^{C_c\times d}$ combines that input with its static-schema
context to produce write features.
The $\pi\tanh(F)$ transformation bounds each coordinate of a position's
phase increment to $[-\pi,\pi]$ before distribution.
The matrix product in Equation~\ref{eq:core-commit} sums these weighted
increments over positions into an $M\times d$ update; it does not average
over chunk length. The bound therefore applies to individual increments,
not the total chunk update.
Updates remain pending while the chunk reads the unchanged $S_c$.
At its boundary, commit adds the accumulated angles and takes the
elementwise remainder modulo $2\pi$, making the new state readable in the
next chunk. This reduction preserves the sine/cosine representation; it
is not a normalization across stored observations.

The layer applies ordinary causal attention to $X+R$ inside the chunk,
without access to previous chunks' K/V.
Its output combines the original input residual $H$, the attended output,
and a learned projection of $F$, followed by a residual MLP.
This local residual uses $F$ before the bounding nonlinearity, providing a
training signal to the write-feature branch even before a later chunk
reads the committed state.
Neither this layer's attention output nor its state read enters its writer:
holding $X$ fixed, $\Delta S_c$ is independent of that layer's $S_c$.
This removes a direct same-layer state--read--write feedback path without
detaching gradients through commits. We introduced this separation to
mitigate gradient instability observed in preliminary experiments with
direct same-layer feedback. Later-chunk losses can still train
earlier writes through the committed state, and lower-layer memory can
affect higher-layer writes through $H$.

\paragraph{Attention-assisted training.}
Full-history (no-commit) mode treats an entire episode as one causal chunk with $S=0$.
Schema projections and the local write-feature residual remain active, so this is not a vanilla Transformer.
Switching to recurrence changes visibility and enables commits without changing parameter shapes.
We use full-history initialization and interleaved training batches.

\paragraph{Resource scaling.}
With $L$ layers, persistent state has $LMd$ elements.
Fixed slot count and chunk length give linear work in sequence length, with sequential chunk dependencies.
Aggregation uses token--slot scores and a matrix product, not a token-wise $[B,T,M,d]$ state trajectory.
Training still retains activations and state history; local attention and pending updates also consume memory.
Implementation details are available in the source code provided in the Supplementary Material.

\section{Controlled State-Maintenance Experiments}
\label{sec:experiments}
\paragraph{Task.}
An episode represents a database with $N=2^a$ addresses and $K=2^v$ values,
where $a,v\in\{3,4,5\}$; \texttt{A3V4}, for example, has eight addresses and sixteen values.
Each address initially has the learned default $b(j)=j\bmod K$, not supplied as an input transaction.
Inputs concatenate address bits, value bits, and a validity bit: one writes a value and zero queries it (NULL).
All models use a biased input projection and predict
$v$ bit logits.
Exact-value accuracy requires all bits to be correct.
Only query positions receive classification supervision; neither targets nor predicted answers are fed back.

\paragraph{Maintenance scenarios.}
Retention writes $N/2$ addresses, waits, then queries all addresses.
Disjoint-write interference writes $N/4$, queries all, writes a disjoint $N/4$, waits, and queries all again.
Overwrite writes $N/2$, queries all, rewrites half of that set, waits, and queries all again.
Every write differs from both the old value and the default. A delay of $D$ consists of $D$ NULL queries to never-written addresses, not an interval with no inputs.
It tests persistence through intervening computation, not pure time decay.
We distinguish retained written values, updated values, and never-written defaults; every final read covers the entire address space.

\paragraph{Models and accounting.}
All networks have three layers and width 32. \sm{} uses 176 slots in each layer, with no separate ordinary-attention bottom layer.
It commits every 96 inputs and at write/read segment boundaries, discarding earlier K/V.
Mamba-3 uses official SISO mixers;
Gated DeltaNet uses the FLA implementation without added softmax attention.
Both baselines scan the episode without state resets.
Table~\ref{tab:resources} matches parameters and persistent elements approximately, not total working memory, bytes, or FLOPs.

\begin{table}[t]\centering
\caption{Resource counts for A3V3. A$a$V$v$ denotes $a$-bit addresses ($2^a$ addresses) and $v$-bit values ($2^v$ possible values). All models have three layers and hidden width 32. Across A$a$V$v$ settings, only the input projection $(a+v+1)\to32$ and output bit classifier $32\to v$ change dimensions; backbone dimensions and persistent-state sizes remain fixed. The extra input bit indicates write versus NULL query. Persistent elements exclude temporary buffers and local attention storage. Final target is the final-stage validation exact-value accuracy threshold, required for every evaluated semantic group on two consecutive checks; it is not the long-delay evaluation accuracy.}
\label{tab:resources}\begin{tabular}{lrrr}\toprule
Model & Parameters & Persistent elements & Final target \\\midrule
SchemaMem & 54,336 & 16,896 & 95\% \\
Mamba-3 & 55,128 & 16,752 & 99\% \\
Gated DeltaNet & 55,260 & 17,280 & 99\% \\
\bottomrule\end{tabular}\end{table}

\paragraph{Training.}
\label{sec:training}
A transaction curriculum increases the queried subset from $N/4$ to $N/2$ to $N$, updating half before querying.
Maintenance training then increases the maximum delay through 96, 192, and 384. \sm{} first learns the transaction curriculum in full-history mode, then alternates full-history and recurrent batches 1:1; maintenance uses one full-history batch in four.
Baselines train recurrently throughout.
All use bitwise binary cross-entropy; maintenance balances semantic groups and downweights filler queries.
Training terminates when the final-stage validation accuracy reaches at least 95\% for \sm{} or 99\% for the baselines in two consecutive evaluations.
This is a higher-mastery baseline control, not equal training compute.
A from-scratch reproduction workflow, including curriculum and optimizer settings,
is provided in the source code included in the Supplementary Material.

\paragraph{Computational resources.}
DB training and evaluation used a single NVIDIA GeForce RTX 4090 GPU
(24\,GB VRAM) per run. The software environment used PyTorch 2.10.0
(CUDA 13.0 build), Transformers 5.14.1, Mamba SSM 2.3.2.post1, and
Flash Linear Attention 0.5.2. Table~\ref{tab:cost} reports total optimizer
steps, not wall-clock time or matched computational budgets.

\paragraph{Evaluation.}
We evaluate terminal checkpoints at delays 96, 192, 384, 768, and 1,536, with 512 episodes per scenario and training seeds 42, 43, and 44: nine address/value settings and 81 terminal checkpoints.
All runs are evaluated recurrently, without long-delay checkpoint selection.
All 81 runs reach their respective final targets.
The evaluation generator seed (20261001) is reused
from validation, so these are controlled diagnostics rather than a separately
held-out test set.
Means and sample SDs are over training seeds.
The longest delay is four times the maximum training \emph{delay}, not sequence length.

\section{Results}
\label{sec:results}
\begin{figure}[t]
\centering
\includegraphics[width=\linewidth]{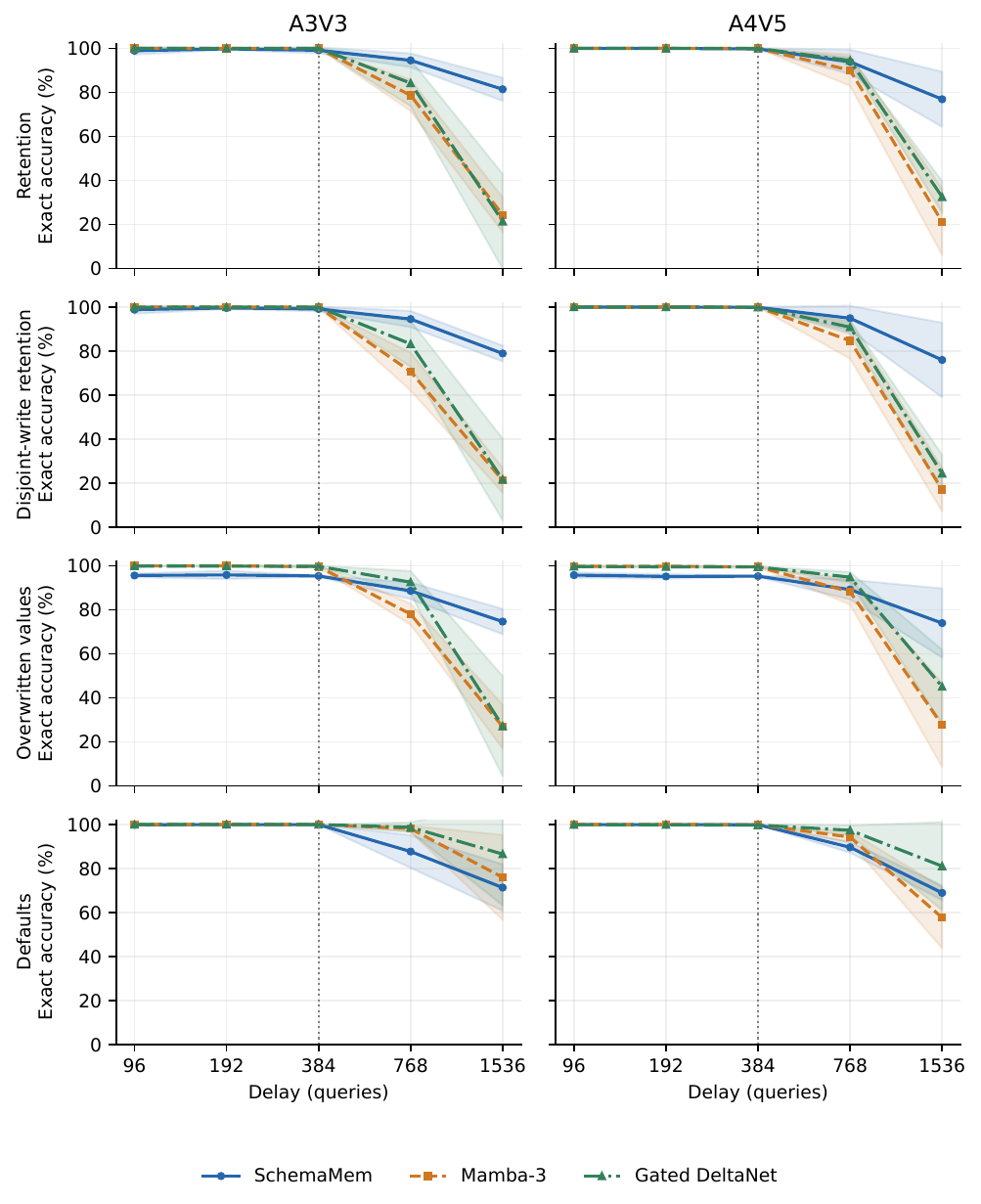}
\caption{Exact-value accuracy versus intervening NULL queries: A3V3 (left)
and A4V5 (right). A$a$V$v$ denotes $a$-bit addresses ($2^a$ addresses) and
$v$-bit values ($2^v$ possible values). Rows separate retention, disjoint-write retention,
overwritten values, and defaults. Lines are three-seed means; shading is
sample SD. The dotted line marks the maximum training delay (384).
Baselines have higher final accuracy targets; \sm{} uses more optimization
steps and full-history training batches. Defaults expose a contrary weakness.}
\label{fig:delay}
\end{figure}

\paragraph{In-range mastery and delayed retention can diverge.}
In A3V3 at delay 384, updated-value accuracy is 95.35\% for \sm{}, 99.58\% for Mamba-3, and 99.74\% for Gated DeltaNet.
At delay 1,536, \sm{} overtakes both baselines: 74.64\%, 26.89\%, and 27.15\%, respectively (Figure~\ref{fig:delay}).
Retained-value accuracy after disjoint writes is 78.91\% for \sm{}, versus 21.26\% and 21.65\%. Poor long-delay performance therefore cannot be explained solely by failure to learn the in-range task.

\begin{table}[t]\centering
\caption{All nine long-delay comparisons: mean exact accuracy (\%) at delay 1,536. A$a$V$v$ denotes $a$-bit addresses ($2^a$ addresses) and $v$-bit values ($2^v$ possible values). SM: SchemaMem; M3: Mamba-3; GDN: Gated DeltaNet. I: retained after disjoint writes; U: overwritten values. Complete means, SDs, default predictions, and per-seed results are in the appendix.}
\label{tab:core-results}\begin{tabular}{lrrrrrr}\toprule
& \multicolumn{3}{c}{Retained-I} & \multicolumn{3}{c}{Updated-U} \\
\cmidrule(lr){2-4}\cmidrule(lr){5-7}
Setting & SM & M3 & GDN & SM & M3 & GDN \\\midrule
A3V3 & 78.9 & 21.3 & 21.6 & 74.6 & 26.9 & 27.1 \\
A3V4 & 71.2 & 12.5 & 30.2 & 66.3 & 20.5 & 34.0 \\
A3V5 & 70.2 & 10.9 & 25.0 & 65.8 & 13.9 & 20.9 \\
A4V3 & 78.1 & 28.8 & 31.5 & 77.7 & 38.7 & 36.4 \\
A4V4 & 79.3 & 12.6 & 33.0 & 75.6 & 20.5 & 42.0 \\
A4V5 & 76.0 & 17.1 & 24.4 & 74.0 & 27.7 & 45.1 \\
A5V3 & 66.1 & 34.0 & 55.9 & 64.3 & 47.4 & 70.4 \\
A5V4 & 65.7 & 13.8 & 62.3 & 64.5 & 23.5 & 75.4 \\
A5V5 & 72.5 & 34.7 & 60.8 & 67.4 & 50.1 & 65.8 \\
\bottomrule\end{tabular}\end{table}

Across all nine settings, \sm{} has higher mean retention accuracy at delay
1,536 than both baselines, with and without disjoint writes.
Updated-value recovery favors \sm{} in nine settings against Mamba-3 and
seven against Gated DeltaNet (Table~\ref{tab:core-results}).
These are descriptive comparisons of three-seed means, not nine independent
significance tests.
Complete curves and individual seeds appear in Appendices~\ref{app:results} and \ref{app:seeds}.

\paragraph{The advantage is conditional.}
Baselines generally have better updated-value accuracy within the training range.
They can also lead at intermediate delays: A3V4 Gated DeltaNet reaches 95.18\% at delay 768, versus 85.90\% for \sm{}.
Defaults often favor the baselines; at delay 1,536 in A3V5, \sm{} obtains 64.78\% compared with 96.57\% for Gated DeltaNet, which outperforms \sm{} on defaults in every grid cell.
At higher address load, \sm{} trails Gated DeltaNet on updated values in A5V3 (64.26\% versus 70.43\%) and A5V4 (64.49\% versus 75.37\%).
A5V5 is close (67.40\% versus 65.80\%), with substantial Gated DeltaNet seed variation.

\paragraph{Dynamic state carries the episode-specific values.}
Zeroing every layer's $S$ after each commit, without changing weights or local
attention, reduces retained- and updated-value accuracy to below 1\% for all
27 \sm{} checkpoints at delays 384 and 1,536.
Default-value predictions largely survive and, at delay 1,536, improve for every
checkpoint. Thus retrieval of written values depends on $S$, while its influence
can also impair default predictions at long delays.
Appendix~\ref{app:state-ablation} gives the paired evaluation and complete tables.

\paragraph{Training cost.}
The countervailing cost is substantial: A3V3 takes 88,667 mean optimizer steps for \sm{}, versus 3,917 for Mamba-3 and 3,750 for Gated DeltaNet (Table~\ref{tab:cost}, appendix).
Counts include attention-assisted batches and are not equal-FLOP budgets. We claim no training-efficiency advantage.

\section{Discussion and Limitations}
\label{sec:discussion}
\paragraph{What this study establishes.}
The completed experiments identify a regime in which an attention-based, schema-indexed recurrent model loses fewer written values after prolonged intervening computation than the tested small Mamba-3 and Gated DeltaNet models.
Near-perfect performance on the trained delays is not sufficient to predict that behavior.
The finding is therefore not only a limitation of the baselines: \sm{} provides
a constructive alternative with a different retention profile at comparable
parameter and persistent-state sizes, albeit at substantially greater training
cost. This is evidence for a useful design trade-off, not interference-free storage.

\paragraph{A controlled diagnostic for recurrent memory.}
The task directly tests the capability that motivates the architecture:
incorporating new information while preserving unrelated information for later
use. Separating retained, overwritten, and default values distinguishes failures
that an aggregate accuracy can conceal. A model can preserve earlier writes yet
struggle to recover replacements, or recover written values while corrupting
the default mapping for untouched addresses. Increasing the intervening query
stream makes these trade-offs visible without introducing language comprehension,
answer generation, or ambiguous ground truth as additional variables.
Such diagnostics can complement sequence-model evaluations by asking not only
whether retrieval succeeds, but which information survives subsequent computation.
Their value does not depend on one architecture winning every condition:
different failure profiles can guide both architectural changes and training curricula.

\paragraph{Why a learned schema may help.}
The static bank supplies a reusable reference representation before any episode-specific update is stored.
Attention can learn compatible write and read coordinates instead of assigning a meaning to an otherwise anonymous state from scratch.
This is a design motivation, not a demonstrated causal explanation:
a learned-bank ablation, fixed-bank control, and matched training schedules would be required to isolate it.
The $S=0$ control confirms that written-value retrieval uses the dynamic state;
it does not establish which design component explains the retention advantage.
Dense weights can still spread writes over many slots, and additive phase composition can still overwrite useful information.

\paragraph{Retention and optimization are separate challenges.}
The higher in-range targets reached by the baselines do not remove their
long-delay deficit in the identified regime. Conversely, \sm{}'s stronger
retention does not make its slow, seed-sensitive optimization negligible.
These observations motivate improving training efficiency while testing whether
the retention profile is preserved. Whether matched training schedules or
alternative curricula would narrow the gap remains open.

\paragraph{Scope of the comparison.}
The principal limitations are small models, three seeds per setting,
substantial optimization-cost differences, training procedures that differ across families, and a narrow family of synthetic episodes.
Persistent-state dimensions omit temporary attention and accumulation storage.
Structured write/read boundaries are supplied by the experiment, and all models benefit from simple deterministic defaults.
The learned addresses are drawn from a finite space, not unseen-address generalization.
Delayed NULL queries are a specific distractor distribution;
arbitrary interleavings, deletions, random defaults, much larger databases, and continuous online streams remain untested here.

\paragraph{Implications for stateful systems.}
An agent that carries goals, constraints, or intermediate findings across
interactions must preserve still-relevant information while incorporating
revisions. Our task isolates an abstract version of that requirement; it does
not measure its contribution to end-to-end agent performance.
The controlled results motivate testing whether the observed retention profile
transfers to such settings, with language processing and external-tool use held
comparable. External records could supply exact details while learned state
maintains information needed to select and interpret those records.
This division of work is a prospective application, not a capability demonstrated here.

\section{Conclusion}
\sm{} combines chunk-local attention with a learned schema reference and an episode-specific phase state.
Static-schema writing and additive commits offer a simple way to retain recurrent memory while keeping attention as the principal read interface and as a training mode.
Controlled experiments show an encouraging long-delay retention regime alongside clear weaknesses in optimization cost, defaults, and some higher-load conditions.
By exposing distinct preservation and revision failures and evaluating a
concrete alternative, the study supplies a controlled basis for further work
on recurrent state maintenance.
The next step is to isolate which architectural and training choices produce this trade-off, rather than to extrapolate it into a universal memory claim.
These findings also motivate evaluating \sm{} in agent tasks to test whether
its retention advantage supports maintaining and revising information across
extended interactions.

\label{main:lastpage}
\clearpage
\subsection*{AI use statement}
Generative AI tools assisted discussions of the
conceptual framework and hypotheses, experimental design, implementation
and debugging of the models and programmatic synthetic-data generators,
training and evaluation workflows, interpretation of results, literature
search, and manuscript preparation. They also assisted mathematical
notation, figure and table generation, and English-language editing.
The address--value episodes and their labels are generated by explicit
programmed rules, not by language-model responses; the reported numerical
results come from executed training and evaluation runs. Checks of
AI-assisted artifacts include code inspection, executable tests, and
cross-checking paper tables against saved evaluation records. The authors
take responsibility for the final text, claims, code, and results.

\subsection*{Ethics statement}
The experiments reported here use programmatically generated address--value
sequences, without human participants or personal data. They test a narrow
state-maintenance capability rather than the safety or reliability of an
autonomous agent. Applying persistent memory to human-facing systems would
add privacy, consent, and deletion requirements that this synthetic study
does not evaluate.

\subsection*{Reproducibility statement}
Section~\ref{sec:method} specifies the memory update and attention modes,
and Section~\ref{sec:experiments} describes the task, model comparison,
curriculum, and evaluation protocol. Appendices~\ref{app:results}
and~\ref{app:seeds} report the complete experimental grid, training costs,
and individual-seed outcomes. Appendix~\ref{app:state-ablation} specifies the
paired dynamic-state ablation. Reported tables and
figures were generated from a frozen numerical snapshot with source and
checkpoint hashes. The Supplementary Material provides the model implementation,
task generators, configurations, standalone training and evaluation scripts,
plotting functionality, and setup instructions. Trained checkpoints are not
included; the supplied workflow trains models from scratch and evaluates the
resulting checkpoints.

\clearpage
\bibliographystyle{iclr2027_conference}
\bibliography{ref}
\clearpage
\appendix
\section{Extended Results}
\label{app:results}
\subsection{Complete long-delay comparisons}
Table~\ref{tab:longdelay} expands the main comparison to all four reported
prediction groups across nine evaluated settings (81 checkpoints).
Hidden width, slot count, depth, and baseline state sizes remain fixed as
address and value widths vary. Exact-value accuracy requires every value
bit to be correct, so increasing value width makes this criterion more
demanding even at a fixed bit accuracy.

\begin{table}[htp]\centering
\caption{Delay 1,536: exact-value accuracy (\%, mean $\pm$ sample SD, three seeds). A$a$V$v$ denotes $a$-bit addresses ($2^a$ addresses) and $v$-bit values ($2^v$ possible values). SM: SchemaMem; M3: Mamba-3; GDN: Gated DeltaNet. Retained-R is the retention-only group; Retained-I is the retained group after disjoint writes; Updated and Default are from the overwrite scenario. All 81 endpoints are included.}
\label{tab:longdelay}\begin{tabular}{llrrrr}\toprule
Setting & Model & Retained-R & Retained-I & Updated & Default \\\midrule
A3V3 & SM & $81.4\,\pm\,5.3$ & $78.9\,\pm\,3.6$ & $74.6\,\pm\,5.8$ & $71.3\,\pm\,10.6$ \\
A3V3 & M3 & $24.3\,\pm\,7.9$ & $21.3\,\pm\,5.6$ & $26.9\,\pm\,10.0$ & $76.0\,\pm\,19.3$ \\
A3V3 & GDN & $21.4\,\pm\,21.6$ & $21.6\,\pm\,18.6$ & $27.1\,\pm\,22.7$ & $86.5\,\pm\,23.3$ \\
\midrule
A3V4 & SM & $71.9\,\pm\,9.3$ & $71.2\,\pm\,9.1$ & $66.3\,\pm\,8.8$ & $71.7\,\pm\,13.5$ \\
A3V4 & M3 & $15.0\,\pm\,3.7$ & $12.5\,\pm\,3.7$ & $20.5\,\pm\,2.5$ & $90.3\,\pm\,3.8$ \\
A3V4 & GDN & $29.7\,\pm\,8.7$ & $30.2\,\pm\,5.8$ & $34.0\,\pm\,8.3$ & $91.3\,\pm\,6.1$ \\
\midrule
A3V5 & SM & $72.5\,\pm\,25.3$ & $70.2\,\pm\,27.5$ & $65.8\,\pm\,21.8$ & $64.8\,\pm\,21.7$ \\
A3V5 & M3 & $12.7\,\pm\,11.3$ & $10.9\,\pm\,8.7$ & $13.9\,\pm\,11.8$ & $54.8\,\pm\,5.8$ \\
A3V5 & GDN & $21.9\,\pm\,16.1$ & $25.0\,\pm\,20.5$ & $20.9\,\pm\,14.8$ & $96.6\,\pm\,5.2$ \\
\midrule
A4V3 & SM & $80.2\,\pm\,24.2$ & $78.1\,\pm\,24.8$ & $77.7\,\pm\,18.3$ & $62.8\,\pm\,14.3$ \\
A4V3 & M3 & $34.4\,\pm\,11.4$ & $28.8\,\pm\,9.9$ & $38.7\,\pm\,9.4$ & $78.7\,\pm\,9.3$ \\
A4V3 & GDN & $37.1\,\pm\,41.0$ & $31.5\,\pm\,37.3$ & $36.4\,\pm\,39.6$ & $72.0\,\pm\,22.9$ \\
\midrule
A4V4 & SM & $82.0\,\pm\,6.4$ & $79.3\,\pm\,5.9$ & $75.6\,\pm\,1.8$ & $62.1\,\pm\,15.3$ \\
A4V4 & M3 & $15.4\,\pm\,5.1$ & $12.6\,\pm\,4.2$ & $20.5\,\pm\,9.2$ & $62.4\,\pm\,8.5$ \\
A4V4 & GDN & $43.3\,\pm\,24.1$ & $33.0\,\pm\,19.2$ & $42.0\,\pm\,18.1$ & $70.3\,\pm\,29.4$ \\
\midrule
A4V5 & SM & $76.9\,\pm\,12.5$ & $76.0\,\pm\,16.9$ & $74.0\,\pm\,15.7$ & $68.9\,\pm\,3.3$ \\
A4V5 & M3 & $21.2\,\pm\,15.0$ & $17.1\,\pm\,9.9$ & $27.7\,\pm\,19.1$ & $57.7\,\pm\,13.9$ \\
A4V5 & GDN & $32.5\,\pm\,7.2$ & $24.4\,\pm\,8.5$ & $45.1\,\pm\,16.8$ & $81.1\,\pm\,20.0$ \\
\midrule
A5V3 & SM & $69.5\,\pm\,8.9$ & $66.1\,\pm\,9.4$ & $64.3\,\pm\,12.4$ & $59.4\,\pm\,2.1$ \\
A5V3 & M3 & $42.7\,\pm\,5.7$ & $34.0\,\pm\,5.4$ & $47.4\,\pm\,7.0$ & $70.5\,\pm\,8.5$ \\
A5V3 & GDN & $64.2\,\pm\,21.1$ & $55.9\,\pm\,20.8$ & $70.4\,\pm\,22.6$ & $73.1\,\pm\,10.4$ \\
\midrule
A5V4 & SM & $68.7\,\pm\,11.8$ & $65.7\,\pm\,12.0$ & $64.5\,\pm\,10.3$ & $59.5\,\pm\,2.5$ \\
A5V4 & M3 & $17.2\,\pm\,2.2$ & $13.8\,\pm\,1.1$ & $23.5\,\pm\,4.5$ & $57.2\,\pm\,18.1$ \\
A5V4 & GDN & $68.1\,\pm\,30.8$ & $62.3\,\pm\,30.7$ & $75.4\,\pm\,24.8$ & $90.8\,\pm\,4.1$ \\
\midrule
A5V5 & SM & $72.3\,\pm\,10.6$ & $72.5\,\pm\,7.6$ & $67.4\,\pm\,4.3$ & $56.9\,\pm\,8.3$ \\
A5V5 & M3 & $46.3\,\pm\,13.1$ & $34.7\,\pm\,11.8$ & $50.1\,\pm\,8.7$ & $69.0\,\pm\,4.2$ \\
A5V5 & GDN & $64.9\,\pm\,32.1$ & $60.8\,\pm\,32.0$ & $65.8\,\pm\,26.6$ & $84.7\,\pm\,11.5$ \\
\bottomrule\end{tabular}\end{table}

The mean retention advantage does not imply superiority on every seed or
prediction group. For example, A4V3 updated-value accuracies for seeds
42/43/44 are 57.03/84.23/91.85\% for \sm{} and 79.10/29.20/0.93\% for
Gated DeltaNet. Gated DeltaNet outperforms \sm{} on updated values in A5V3
and A5V4, and on defaults in all nine settings. In A5V5, the updated-value means
are close (67.40 $\pm$ 4.31\% versus 65.80 $\pm$ 26.62\%); this is not
strong evidence of an advantage.


\clearpage
\subsection{Training cost}

\begin{table}[htp]\centering
\caption{Total optimizer steps (transaction plus maintenance), mean over three seeds. A$a$V$v$ denotes $a$-bit addresses ($2^a$ addresses) and $v$-bit values ($2^v$ possible values). SchemaMem totals include full-history training; these are not matched-FLOP budgets.}
\label{tab:cost}\begin{tabular}{lrrr}\toprule
Setting & SchemaMem & Mamba-3 & Gated DeltaNet \\\midrule
A3V3 & 88,667 & 3,917 & 3,750 \\
A3V4 & 58,000 & 4,083 & 4,333 \\
A3V5 & 58,417 & 3,833 & 5,583 \\
A4V3 & 65,667 & 4,917 & 6,667 \\
A4V4 & 36,917 & 4,833 & 6,000 \\
A4V5 & 63,000 & 5,083 & 5,917 \\
A5V3 & 40,833 & 6,583 & 11,167 \\
A5V4 & 36,917 & 7,083 & 29,750 \\
A5V5 & 35,667 & 8,750 & 12,333 \\
\bottomrule\end{tabular}\end{table}

Table~\ref{tab:cost} reports total optimizer steps, including \sm{}'s
full-history batches. \sm{} requires substantially more training, with
large differences between seeds (Appendix~\ref{app:seeds}). These are not
equal-FLOP or equal-example-length budgets and do not support a
training-efficiency advantage. Architecture and attention-assisted training
are not isolated by this comparison.

\clearpage
\subsection{Complete delay profiles and within-seed degradation}
\begin{figure}[htp]
\centering\includegraphics[width=\linewidth]{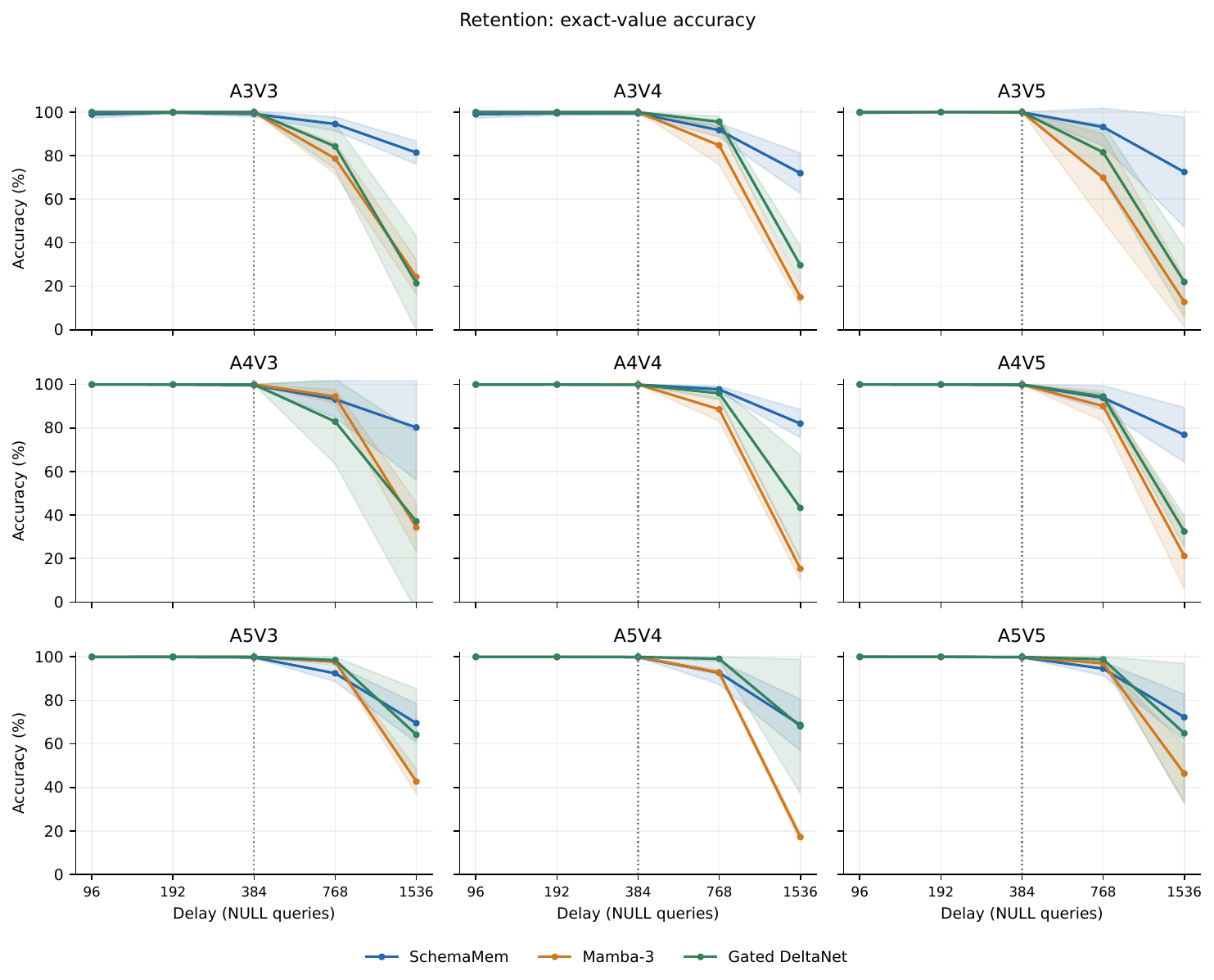}
\caption{Retention-only exact accuracy across the complete grid.
A$a$V$v$ denotes $a$-bit addresses ($2^a$ addresses) and $v$-bit values ($2^v$ possible values).
Shading is
three-seed sample SD, not a confidence interval. The dotted line marks the
training maximum delay (384); delay is measured in NULL queries. All 81 runs
are included.}
\label{fig:all-retention}
\end{figure}

\begin{figure}[p]
\centering\includegraphics[width=\linewidth]{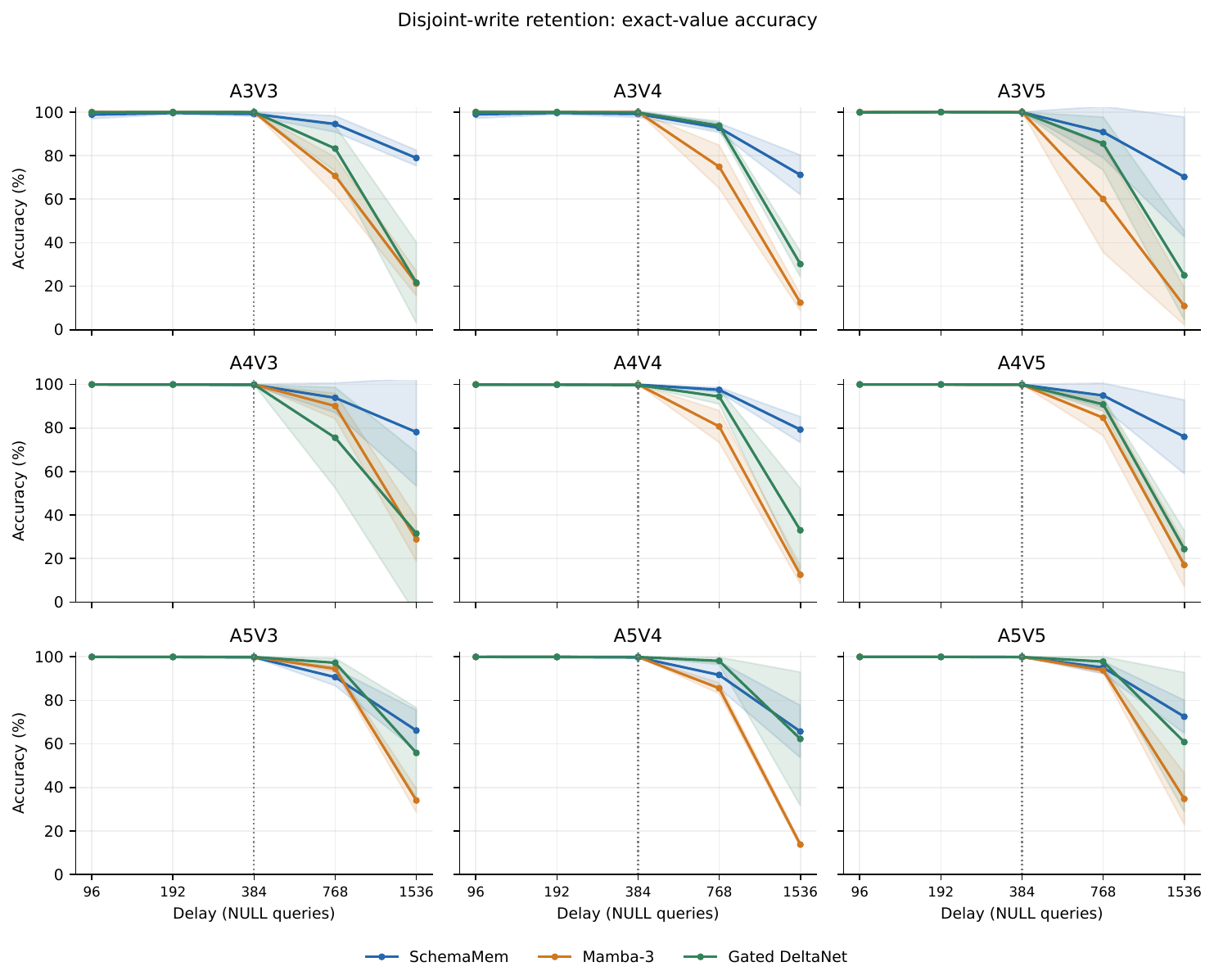}
\caption{Retained written values after disjoint writes, with the same
protocol and sample-SD convention as Figure~\ref{fig:all-retention}. The
longest-delay mean advantage is present in all nine cells, but individual
seeds and shorter delays do not uniformly favor \sm{}.
A$a$V$v$ denotes $a$-bit addresses ($2^a$ addresses) and $v$-bit values ($2^v$ possible values).}
\label{fig:all-interference}
\end{figure}

\begin{figure}[p]
\centering\includegraphics[width=\linewidth]{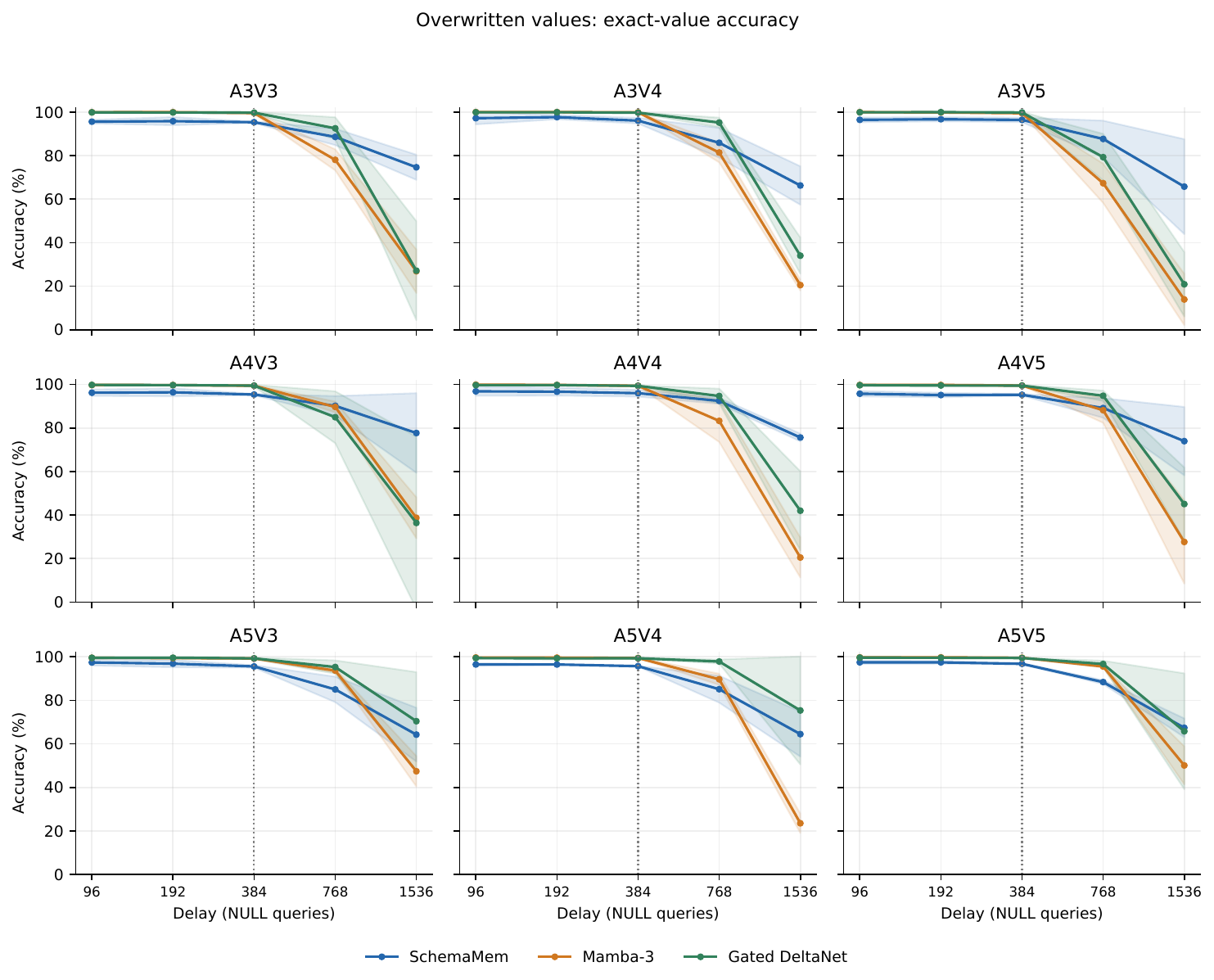}
\caption{Recovery of overwritten new values across the complete grid.
A$a$V$v$ denotes $a$-bit addresses ($2^a$ addresses) and $v$-bit values ($2^v$ possible values).
Shading shows sample SD. The Gated DeltaNet advantage in A5V3/A5V4 and its
near tie in A5V5 at delay 1,536 delimit the claim.}
\label{fig:all-updated}
\end{figure}

\begin{figure}[p]
\centering\includegraphics[width=\linewidth]{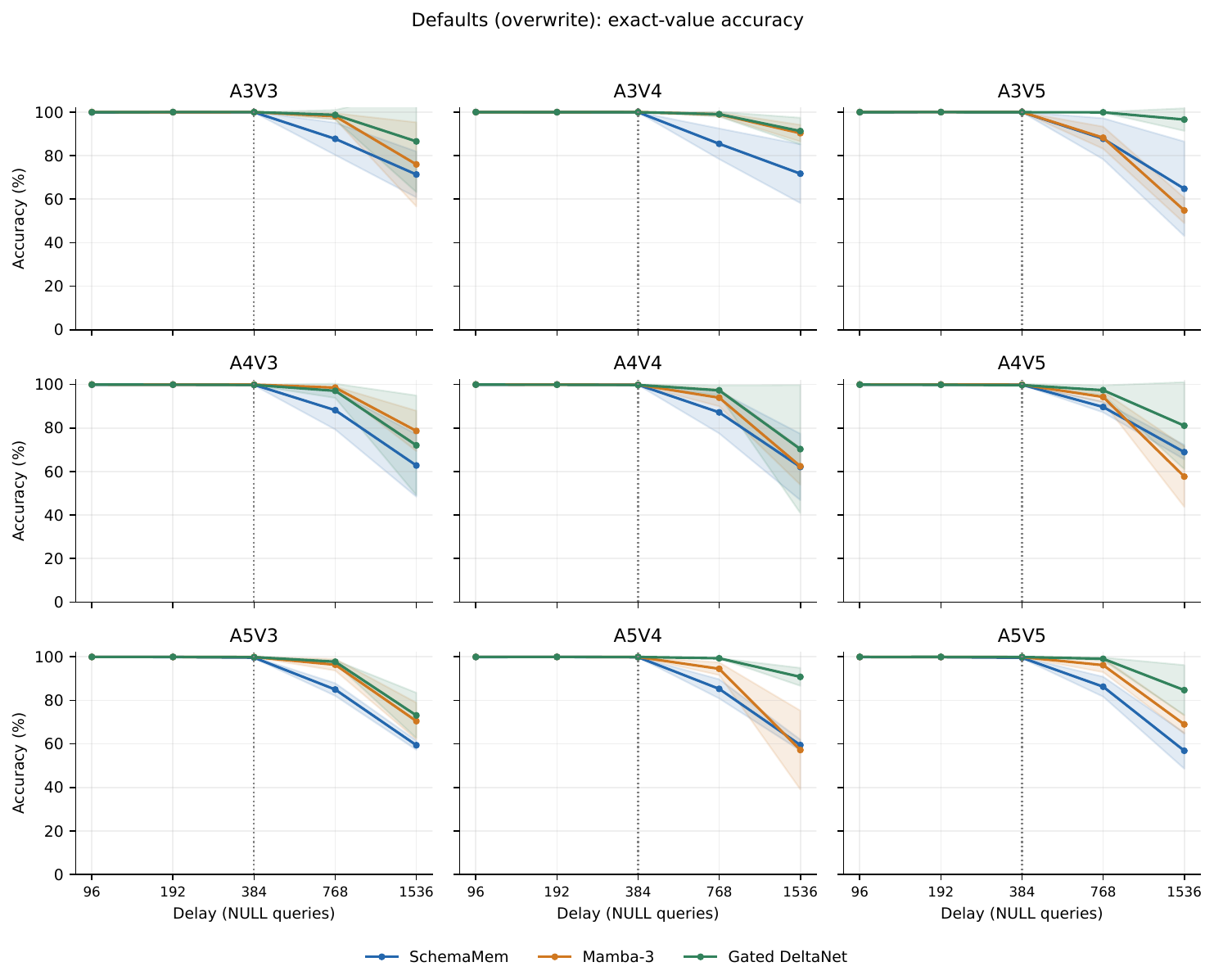}
\caption{Never-written default values in overwrite episodes. Shading shows
sample SD. Gated DeltaNet outperforms \sm{} at delay 1,536 in every cell; written-value
retention should not be interpreted as preserving every useful computation.
A$a$V$v$ denotes $a$-bit addresses ($2^a$ addresses) and $v$-bit values ($2^v$ possible values).}
\label{fig:all-default}
\end{figure}

\begin{figure}[p]
\centering\includegraphics[width=\linewidth]{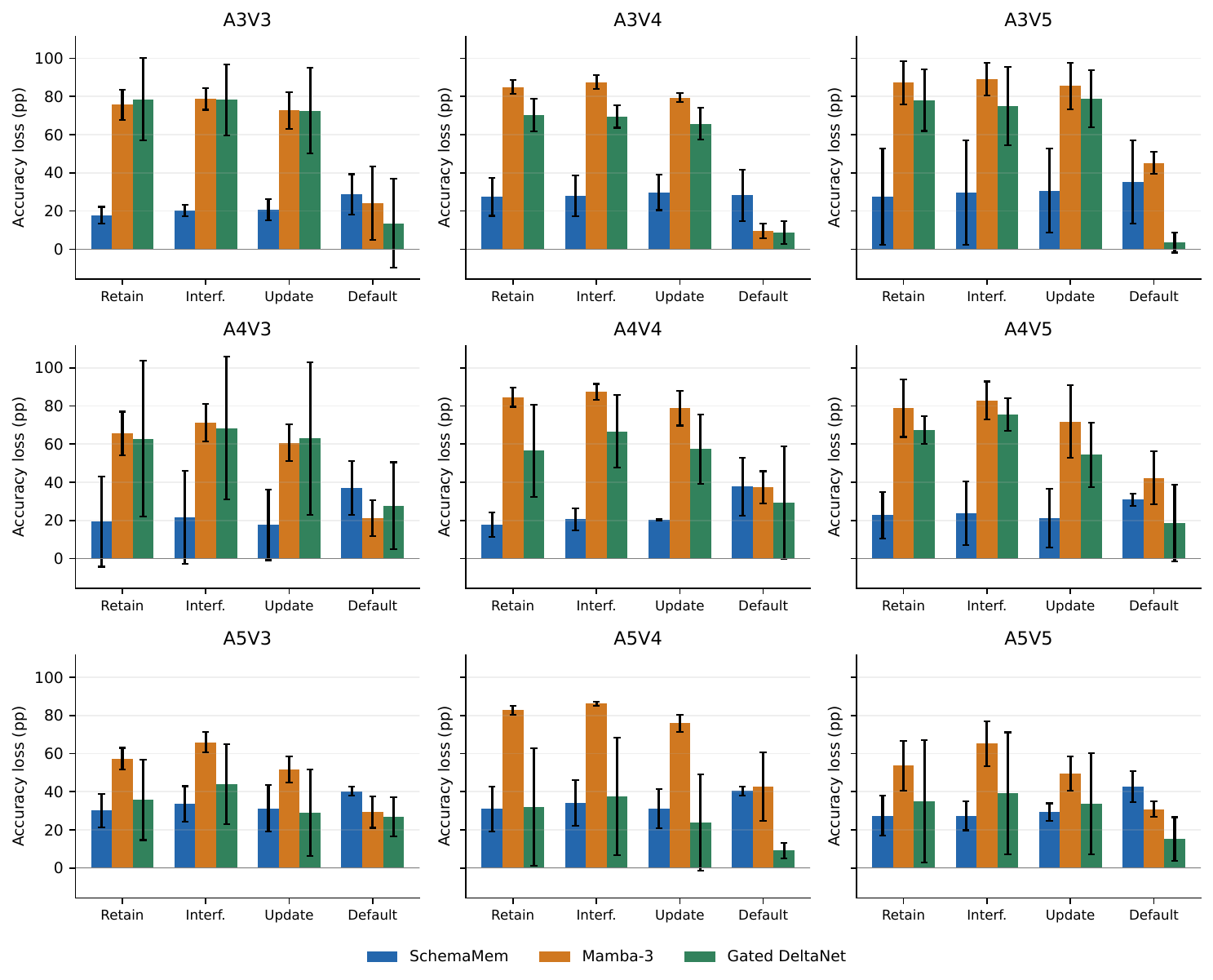}
\caption{Exact-accuracy degradation from delay 384 to 1,536 (percentage
points; lower is better). Differences are formed within each training seed,
then summarized by mean $\pm$ sample SD. This is not a causal estimate of
forgetting or a matched-accuracy comparison. All three seeds are included.
A$a$V$v$ denotes $a$-bit addresses ($2^a$ addresses) and $v$-bit values ($2^v$ possible values).}
\label{fig:drops}
\end{figure}

Figures~\ref{fig:all-retention}--\ref{fig:all-default} show all nine settings
and all evaluated delays, including regimes where baselines lead.
Figure~\ref{fig:drops} reports each seed's accuracy decrease from delay 384
to 1,536, then averages these differences with sample SD. A smaller drop
need not imply higher absolute accuracy, so both views are retained.
The evaluation episodes and checkpoint-selection policy are specified in
Section~\ref{sec:experiments}.

\clearpage
\section{Individual Seeds and Complete Diagnostic Results}
\label{app:seeds}
\subsection{Seed-level variation}
\begin{figure}[p]
\centering
\includegraphics[width=.92\linewidth,height=.85\textheight,keepaspectratio]{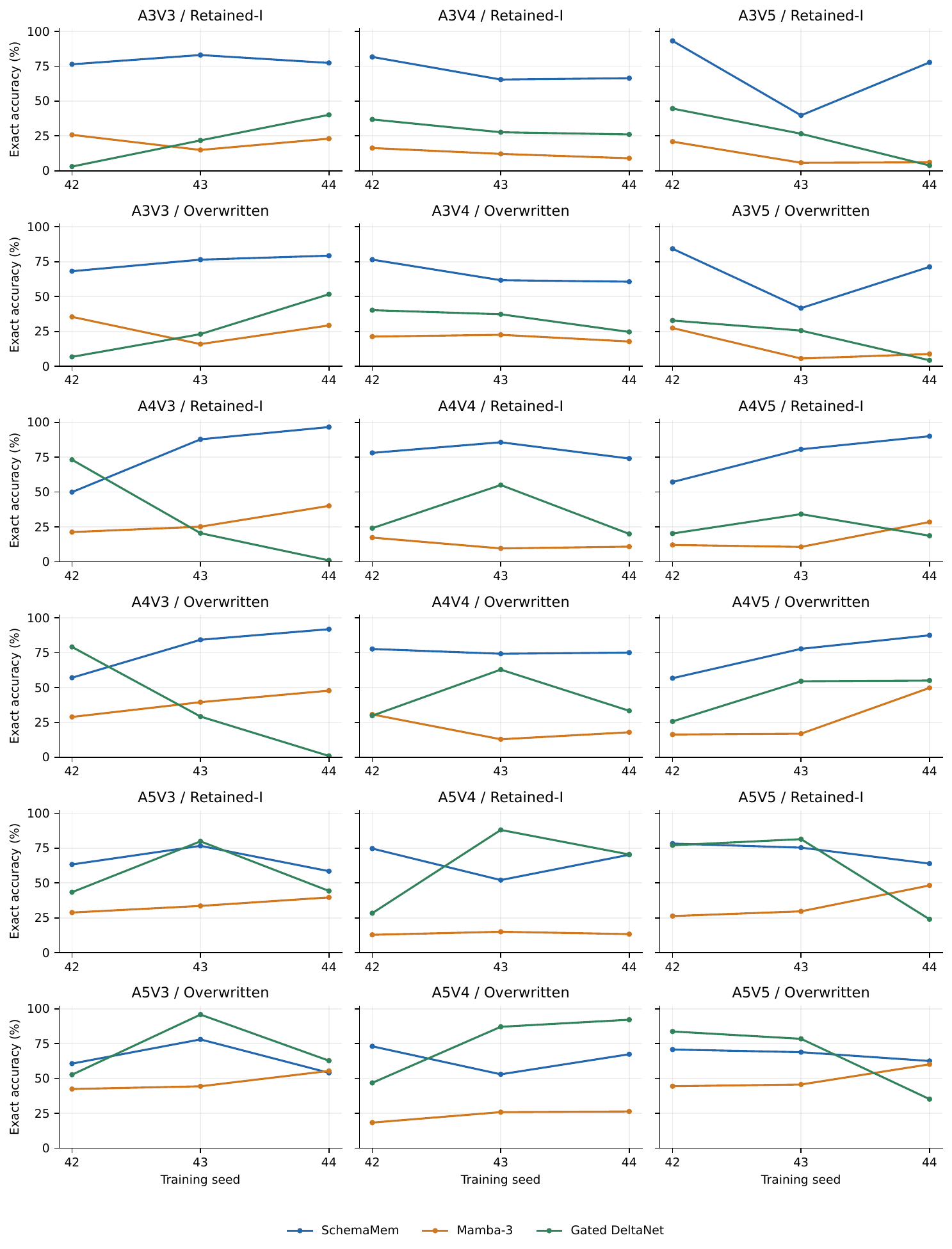}
\caption{Individual training seeds for all nine settings, at delay 1,536.
A$a$V$v$ denotes $a$-bit addresses ($2^a$ addresses) and $v$-bit values ($2^v$ possible values).
Each address-width block has two rows: retained values after disjoint writes,
then overwritten new values. Points are not independently replicated datasets. Lines connect
training seeds only as a visual guide. Large seed effects are retained in
the reported means and sample standard deviations.}
\label{fig:seeds}
\end{figure}

Figure~\ref{fig:seeds} shows individual-seed long-delay results across the
address/value grid. Table~\ref{tab:seeds} lists all 81 terminal checkpoints,
their training costs, and completion status. All runs reach their respective
final targets; no seed is removed from the aggregate results.

\begingroup
\begin{longtable}{llrrrrrr}
\caption{Per-seed endpoints. A$a$V$v$ denotes $a$-bit addresses ($2^a$ addresses) and $v$-bit values ($2^v$ possible values). SM: SchemaMem; M3: Mamba-3; GDN: Gated DeltaNet. Tx/Maint: transaction/maintenance optimizer steps. I/U: exact accuracy at delay 1,536 for retained interference/overwritten values. Pass refers to the configured final target, not to long-delay accuracy.}\label{tab:seeds}\\
\toprule Setting & Model & Seed & Tx & Maint & I (\%) & U (\%) & Pass \\\midrule\endfirsthead
\toprule Setting & Model & Seed & Tx & Maint & I (\%) & U (\%) & Pass \\\midrule\endhead
A3V3 & SM & 42 & 8,750 & 42,250 & 76.37 & 68.16 & yes \\
A3V3 & SM & 43 & 14,500 & 97,000 & 83.01 & 76.46 & yes \\
A3V3 & SM & 44 & 18,000 & 85,500 & 77.34 & 79.30 & yes \\
A3V3 & M3 & 42 & 1,500 & 2,000 & 25.78 & 35.45 & yes \\
A3V3 & M3 & 43 & 1,500 & 2,250 & 14.94 & 15.92 & yes \\
A3V3 & M3 & 44 & 1,500 & 3,000 & 23.05 & 29.30 & yes \\
A3V3 & GDN & 42 & 1,750 & 1,750 & 3.03 & 6.74 & yes \\
A3V3 & GDN & 43 & 1,500 & 2,750 & 21.78 & 23.05 & yes \\
A3V3 & GDN & 44 & 1,750 & 1,750 & 40.14 & 51.66 & yes \\
\midrule
A3V4 & SM & 42 & 9,250 & 31,750 & 81.64 & 76.46 & yes \\
A3V4 & SM & 43 & 6,000 & 32,250 & 65.43 & 61.72 & yes \\
A3V4 & SM & 44 & 16,500 & 78,250 & 66.41 & 60.64 & yes \\
A3V4 & M3 & 42 & 1,500 & 2,500 & 16.31 & 21.29 & yes \\
A3V4 & M3 & 43 & 1,500 & 2,500 & 12.11 & 22.56 & yes \\
A3V4 & M3 & 44 & 1,500 & 2,750 & 8.98 & 17.77 & yes \\
A3V4 & GDN & 42 & 1,500 & 3,500 & 36.82 & 40.23 & yes \\
A3V4 & GDN & 43 & 1,500 & 2,250 & 27.64 & 37.30 & yes \\
A3V4 & GDN & 44 & 1,500 & 2,750 & 26.07 & 24.61 & yes \\
\midrule
A3V5 & SM & 42 & 12,000 & 30,500 & 93.16 & 84.28 & yes \\
A3V5 & SM & 43 & 16,000 & 53,000 & 39.75 & 41.70 & yes \\
A3V5 & SM & 44 & 16,750 & 47,000 & 77.73 & 71.29 & yes \\
A3V5 & M3 & 42 & 1,500 & 2,500 & 20.90 & 27.44 & yes \\
A3V5 & M3 & 43 & 1,500 & 2,250 & 5.76 & 5.57 & yes \\
A3V5 & M3 & 44 & 1,500 & 2,250 & 6.05 & 8.79 & yes \\
A3V5 & GDN & 42 & 2,500 & 4,000 & 44.63 & 32.81 & yes \\
A3V5 & GDN & 43 & 1,500 & 5,000 & 26.56 & 25.59 & yes \\
A3V5 & GDN & 44 & 1,500 & 2,250 & 3.81 & 4.30 & yes \\
\midrule
A4V3 & SM & 42 & 13,750 & 43,500 & 49.95 & 57.03 & yes \\
A4V3 & SM & 43 & 13,000 & 34,250 & 87.79 & 84.23 & yes \\
A4V3 & SM & 44 & 9,500 & 83,000 & 96.58 & 91.85 & yes \\
A4V3 & M3 & 42 & 1,500 & 3,000 & 21.29 & 28.91 & yes \\
A4V3 & M3 & 43 & 1,500 & 3,250 & 25.15 & 39.50 & yes \\
A4V3 & M3 & 44 & 1,500 & 4,000 & 40.09 & 47.75 & yes \\
A4V3 & GDN & 42 & 2,750 & 8,250 & 73.14 & 79.10 & yes \\
A4V3 & GDN & 43 & 2,500 & 2,750 & 20.51 & 29.20 & yes \\
A4V3 & GDN & 44 & 1,750 & 2,000 & 0.98 & 0.93 & yes \\
\midrule
A4V4 & SM & 42 & 7,500 & 29,250 & 78.08 & 77.64 & yes \\
A4V4 & SM & 43 & 7,000 & 19,500 & 85.69 & 74.22 & yes \\
A4V4 & SM & 44 & 11,750 & 35,750 & 74.02 & 75.05 & yes \\
A4V4 & M3 & 42 & 1,500 & 3,000 & 17.33 & 30.71 & yes \\
A4V4 & M3 & 43 & 1,500 & 2,500 & 9.57 & 12.89 & yes \\
A4V4 & M3 & 44 & 1,500 & 4,500 & 10.84 & 17.92 & yes \\
A4V4 & GDN & 42 & 2,000 & 3,750 & 24.07 & 29.83 & yes \\
A4V4 & GDN & 43 & 2,500 & 4,000 & 55.03 & 62.84 & yes \\
A4V4 & GDN & 44 & 2,250 & 3,500 & 19.97 & 33.30 & yes \\
\midrule
A4V5 & SM & 42 & 6,750 & 23,500 & 57.18 & 56.69 & yes \\
A4V5 & SM & 43 & 18,750 & 91,000 & 80.66 & 77.73 & yes \\
A4V5 & SM & 44 & 8,750 & 40,250 & 90.04 & 87.45 & yes \\
A4V5 & M3 & 42 & 2,000 & 4,500 & 12.06 & 16.31 & yes \\
A4V5 & M3 & 43 & 1,500 & 2,500 & 10.64 & 16.89 & yes \\
A4V5 & M3 & 44 & 1,750 & 3,000 & 28.52 & 49.76 & yes \\
A4V5 & GDN & 42 & 2,000 & 1,750 & 20.26 & 25.68 & yes \\
A4V5 & GDN & 43 & 2,750 & 2,000 & 34.18 & 54.49 & yes \\
A4V5 & GDN & 44 & 3,250 & 6,000 & 18.65 & 55.03 & yes \\
\midrule
A5V3 & SM & 42 & 12,250 & 17,750 & 63.31 & 60.67 & yes \\
A5V3 & SM & 43 & 14,250 & 41,000 & 76.66 & 78.03 & yes \\
A5V3 & SM & 44 & 12,250 & 25,000 & 58.47 & 54.08 & yes \\
A5V3 & M3 & 42 & 2,000 & 4,750 & 28.83 & 42.46 & yes \\
A5V3 & M3 & 43 & 2,000 & 3,500 & 33.57 & 44.41 & yes \\
A5V3 & M3 & 44 & 2,000 & 5,500 & 39.67 & 55.42 & yes \\
A5V3 & GDN & 42 & 2,750 & 4,750 & 43.41 & 52.71 & yes \\
A5V3 & GDN & 43 & 2,750 & 9,750 & 79.88 & 95.83 & yes \\
A5V3 & GDN & 44 & 7,750 & 5,750 & 44.31 & 62.77 & yes \\
\midrule
A5V4 & SM & 42 & 14,500 & 41,500 & 74.71 & 73.07 & yes \\
A5V4 & SM & 43 & 10,500 & 9,750 & 52.10 & 53.00 & yes \\
A5V4 & SM & 44 & 11,750 & 22,750 & 70.29 & 67.41 & yes \\
A5V4 & M3 & 42 & 2,000 & 5,500 & 12.87 & 18.38 & yes \\
A5V4 & M3 & 43 & 1,750 & 4,000 & 15.04 & 25.88 & yes \\
A5V4 & M3 & 44 & 2,000 & 6,000 & 13.35 & 26.37 & yes \\
A5V4 & GDN & 42 & 3,750 & 4,250 & 28.37 & 46.88 & yes \\
A5V4 & GDN & 43 & 15,000 & 58,750 & 88.06 & 87.08 & yes \\
A5V4 & GDN & 44 & 3,000 & 4,500 & 70.46 & 92.14 & yes \\
\midrule
A5V5 & SM & 42 & 17,250 & 25,000 & 78.25 & 70.78 & yes \\
A5V5 & SM & 43 & 9,500 & 20,500 & 75.37 & 68.87 & yes \\
A5V5 & SM & 44 & 16,750 & 18,000 & 63.92 & 62.55 & yes \\
A5V5 & M3 & 42 & 4,750 & 6,250 & 26.29 & 44.46 & yes \\
A5V5 & M3 & 43 & 2,750 & 4,750 & 29.69 & 45.70 & yes \\
A5V5 & M3 & 44 & 2,750 & 5,000 & 48.27 & 60.18 & yes \\
A5V5 & GDN & 42 & 3,500 & 3,250 & 77.08 & 83.72 & yes \\
A5V5 & GDN & 43 & 5,500 & 10,000 & 81.45 & 78.47 & yes \\
A5V5 & GDN & 44 & 6,250 & 8,500 & 24.00 & 35.21 & yes \\
\midrule
\end{longtable}\endgroup

\clearpage

\clearpage
\subsection{Scenario-level results}
Table~\ref{tab:groups} reports every semantic group in each scenario at
delay 1,536. These are final-query accuracies, averaged over the three
training seeds with sample SD; intermediate and delay-filler queries are
not included in the reported scores.
\begingroup
\begin{longtable}{lllrrr}
\caption{All maintenance scenario groups at delay 1,536: exact-value accuracy, mean $\pm$ sample SD (\%). A$a$V$v$ denotes $a$-bit addresses ($2^a$ addresses) and $v$-bit values ($2^v$ possible values). SM: SchemaMem; M3: Mamba-3; GDN: Gated DeltaNet. A dash means the group is absent by construction.}\label{tab:groups}\\
\toprule Setting & Model & Scenario & Retained & Updated & Default \\\midrule\endfirsthead
\toprule Setting & Model & Scenario & Retained & Updated & Default \\\midrule\endhead
A3V3 & SM & retention & $81.4\,\pm\,5.3$ & --- & $71.7\,\pm\,10.2$ \\
A3V3 & SM & interference & $78.9\,\pm\,3.6$ & $84.8\,\pm\,4.6$ & $71.1\,\pm\,10.8$ \\
A3V3 & SM & overwrite & $80.2\,\pm\,1.4$ & $74.6\,\pm\,5.8$ & $71.3\,\pm\,10.6$ \\
A3V3 & M3 & retention & $24.3\,\pm\,7.9$ & --- & $75.1\,\pm\,20.1$ \\
A3V3 & M3 & interference & $21.3\,\pm\,5.6$ & $28.9\,\pm\,10.6$ & $75.3\,\pm\,19.9$ \\
A3V3 & M3 & overwrite & $20.9\,\pm\,6.6$ & $26.9\,\pm\,10.0$ & $76.0\,\pm\,19.3$ \\
A3V3 & GDN & retention & $21.4\,\pm\,21.6$ & --- & $86.3\,\pm\,23.7$ \\
A3V3 & GDN & interference & $21.6\,\pm\,18.6$ & $22.5\,\pm\,24.1$ & $86.2\,\pm\,23.7$ \\
A3V3 & GDN & overwrite & $19.1\,\pm\,16.7$ & $27.1\,\pm\,22.7$ & $86.5\,\pm\,23.3$ \\
\midrule
A3V4 & SM & retention & $71.9\,\pm\,9.3$ & --- & $73.2\,\pm\,12.8$ \\
A3V4 & SM & interference & $71.2\,\pm\,9.1$ & $80.3\,\pm\,11.2$ & $71.3\,\pm\,12.9$ \\
A3V4 & SM & overwrite & $69.6\,\pm\,9.3$ & $66.3\,\pm\,8.8$ & $71.7\,\pm\,13.5$ \\
A3V4 & M3 & retention & $15.0\,\pm\,3.7$ & --- & $90.0\,\pm\,4.9$ \\
A3V4 & M3 & interference & $12.5\,\pm\,3.7$ & $17.2\,\pm\,3.9$ & $89.7\,\pm\,4.3$ \\
A3V4 & M3 & overwrite & $12.2\,\pm\,3.5$ & $20.5\,\pm\,2.5$ & $90.3\,\pm\,3.8$ \\
A3V4 & GDN & retention & $29.7\,\pm\,8.7$ & --- & $91.5\,\pm\,6.3$ \\
A3V4 & GDN & interference & $30.2\,\pm\,5.8$ & $24.8\,\pm\,10.6$ & $89.2\,\pm\,5.9$ \\
A3V4 & GDN & overwrite & $27.3\,\pm\,12.0$ & $34.0\,\pm\,8.3$ & $91.3\,\pm\,6.1$ \\
\midrule
A3V5 & SM & retention & $72.5\,\pm\,25.3$ & --- & $64.9\,\pm\,23.8$ \\
A3V5 & SM & interference & $70.2\,\pm\,27.5$ & $78.5\,\pm\,23.1$ & $66.5\,\pm\,21.4$ \\
A3V5 & SM & overwrite & $71.3\,\pm\,23.3$ & $65.8\,\pm\,21.8$ & $64.8\,\pm\,21.7$ \\
A3V5 & M3 & retention & $12.7\,\pm\,11.3$ & --- & $55.4\,\pm\,6.5$ \\
A3V5 & M3 & interference & $10.9\,\pm\,8.7$ & $16.4\,\pm\,16.7$ & $56.2\,\pm\,6.2$ \\
A3V5 & M3 & overwrite & $11.1\,\pm\,9.0$ & $13.9\,\pm\,11.8$ & $54.8\,\pm\,5.8$ \\
A3V5 & GDN & retention & $21.9\,\pm\,16.1$ & --- & $96.7\,\pm\,4.7$ \\
A3V5 & GDN & interference & $25.0\,\pm\,20.5$ & $20.9\,\pm\,16.0$ & $91.6\,\pm\,13.8$ \\
A3V5 & GDN & overwrite & $24.0\,\pm\,18.2$ & $20.9\,\pm\,14.8$ & $96.6\,\pm\,5.2$ \\
\midrule
A4V3 & SM & retention & $80.2\,\pm\,24.2$ & --- & $64.6\,\pm\,15.4$ \\
A4V3 & SM & interference & $78.1\,\pm\,24.8$ & $86.1\,\pm\,16.2$ & $62.9\,\pm\,14.6$ \\
A4V3 & SM & overwrite & $78.8\,\pm\,24.3$ & $77.7\,\pm\,18.3$ & $62.8\,\pm\,14.3$ \\
A4V3 & M3 & retention & $34.4\,\pm\,11.4$ & --- & $79.2\,\pm\,9.3$ \\
A4V3 & M3 & interference & $28.8\,\pm\,9.9$ & $38.3\,\pm\,9.2$ & $78.9\,\pm\,9.1$ \\
A4V3 & M3 & overwrite & $29.7\,\pm\,9.9$ & $38.7\,\pm\,9.4$ & $78.7\,\pm\,9.3$ \\
A4V3 & GDN & retention & $37.1\,\pm\,41.0$ & --- & $75.3\,\pm\,19.9$ \\
A4V3 & GDN & interference & $31.5\,\pm\,37.3$ & $41.0\,\pm\,45.3$ & $72.4\,\pm\,22.3$ \\
A4V3 & GDN & overwrite & $39.0\,\pm\,42.3$ & $36.4\,\pm\,39.6$ & $72.0\,\pm\,22.9$ \\
\midrule
A4V4 & SM & retention & $82.0\,\pm\,6.4$ & --- & $62.7\,\pm\,14.2$ \\
A4V4 & SM & interference & $79.3\,\pm\,5.9$ & $91.5\,\pm\,2.0$ & $62.5\,\pm\,15.1$ \\
A4V4 & SM & overwrite & $80.6\,\pm\,6.3$ & $75.6\,\pm\,1.8$ & $62.1\,\pm\,15.3$ \\
A4V4 & M3 & retention & $15.4\,\pm\,5.1$ & --- & $64.5\,\pm\,10.2$ \\
A4V4 & M3 & interference & $12.6\,\pm\,4.2$ & $19.2\,\pm\,8.4$ & $63.6\,\pm\,9.0$ \\
A4V4 & M3 & overwrite & $13.4\,\pm\,4.3$ & $20.5\,\pm\,9.2$ & $62.4\,\pm\,8.5$ \\
A4V4 & GDN & retention & $43.3\,\pm\,24.1$ & --- & $74.1\,\pm\,24.4$ \\
A4V4 & GDN & interference & $33.0\,\pm\,19.2$ & $41.9\,\pm\,22.5$ & $65.4\,\pm\,34.4$ \\
A4V4 & GDN & overwrite & $45.0\,\pm\,25.9$ & $42.0\,\pm\,18.1$ & $70.3\,\pm\,29.4$ \\
\midrule
A4V5 & SM & retention & $76.9\,\pm\,12.5$ & --- & $69.4\,\pm\,2.9$ \\
A4V5 & SM & interference & $76.0\,\pm\,16.9$ & $88.5\,\pm\,6.0$ & $68.7\,\pm\,3.1$ \\
A4V5 & SM & overwrite & $75.9\,\pm\,17.3$ & $74.0\,\pm\,15.7$ & $68.9\,\pm\,3.3$ \\
A4V5 & M3 & retention & $21.2\,\pm\,15.0$ & --- & $60.6\,\pm\,13.0$ \\
A4V5 & M3 & interference & $17.1\,\pm\,9.9$ & $29.6\,\pm\,21.1$ & $59.4\,\pm\,13.1$ \\
A4V5 & M3 & overwrite & $16.8\,\pm\,11.7$ & $27.7\,\pm\,19.1$ & $57.7\,\pm\,13.9$ \\
A4V5 & GDN & retention & $32.5\,\pm\,7.2$ & --- & $81.5\,\pm\,22.3$ \\
A4V5 & GDN & interference & $24.4\,\pm\,8.5$ & $34.0\,\pm\,10.6$ & $80.3\,\pm\,23.7$ \\
A4V5 & GDN & overwrite & $33.7\,\pm\,12.6$ & $45.1\,\pm\,16.8$ & $81.1\,\pm\,20.0$ \\
\midrule
A5V3 & SM & retention & $69.5\,\pm\,8.9$ & --- & $61.2\,\pm\,3.4$ \\
A5V3 & SM & interference & $66.1\,\pm\,9.4$ & $76.5\,\pm\,8.8$ & $59.6\,\pm\,3.5$ \\
A5V3 & SM & overwrite & $67.0\,\pm\,9.6$ & $64.3\,\pm\,12.4$ & $59.4\,\pm\,2.1$ \\
A5V3 & M3 & retention & $42.7\,\pm\,5.7$ & --- & $71.8\,\pm\,7.9$ \\
A5V3 & M3 & interference & $34.0\,\pm\,5.4$ & $49.6\,\pm\,6.2$ & $70.7\,\pm\,8.2$ \\
A5V3 & M3 & overwrite & $35.3\,\pm\,5.8$ & $47.4\,\pm\,7.0$ & $70.5\,\pm\,8.5$ \\
A5V3 & GDN & retention & $64.2\,\pm\,21.1$ & --- & $76.1\,\pm\,9.3$ \\
A5V3 & GDN & interference & $55.9\,\pm\,20.8$ & $71.4\,\pm\,22.5$ & $75.3\,\pm\,10.2$ \\
A5V3 & GDN & overwrite & $58.3\,\pm\,24.7$ & $70.4\,\pm\,22.6$ & $73.1\,\pm\,10.4$ \\
\midrule
A5V4 & SM & retention & $68.7\,\pm\,11.8$ & --- & $60.5\,\pm\,2.5$ \\
A5V4 & SM & interference & $65.7\,\pm\,12.0$ & $78.5\,\pm\,10.5$ & $59.7\,\pm\,2.3$ \\
A5V4 & SM & overwrite & $67.0\,\pm\,12.8$ & $64.5\,\pm\,10.3$ & $59.5\,\pm\,2.5$ \\
A5V4 & M3 & retention & $17.2\,\pm\,2.2$ & --- & $59.2\,\pm\,18.3$ \\
A5V4 & M3 & interference & $13.8\,\pm\,1.1$ & $21.6\,\pm\,2.3$ & $57.9\,\pm\,18.3$ \\
A5V4 & M3 & overwrite & $14.4\,\pm\,2.5$ & $23.5\,\pm\,4.5$ & $57.2\,\pm\,18.1$ \\
A5V4 & GDN & retention & $68.1\,\pm\,30.8$ & --- & $91.4\,\pm\,5.4$ \\
A5V4 & GDN & interference & $62.3\,\pm\,30.7$ & $72.4\,\pm\,31.8$ & $91.2\,\pm\,5.6$ \\
A5V4 & GDN & overwrite & $62.0\,\pm\,30.9$ & $75.4\,\pm\,24.8$ & $90.8\,\pm\,4.1$ \\
\midrule
A5V5 & SM & retention & $72.3\,\pm\,10.6$ & --- & $57.9\,\pm\,9.2$ \\
A5V5 & SM & interference & $72.5\,\pm\,7.6$ & $80.3\,\pm\,3.3$ & $57.2\,\pm\,8.5$ \\
A5V5 & SM & overwrite & $72.0\,\pm\,8.3$ & $67.4\,\pm\,4.3$ & $56.9\,\pm\,8.3$ \\
A5V5 & M3 & retention & $46.3\,\pm\,13.1$ & --- & $73.1\,\pm\,2.5$ \\
A5V5 & M3 & interference & $34.7\,\pm\,11.8$ & $56.1\,\pm\,13.3$ & $72.6\,\pm\,2.5$ \\
A5V5 & M3 & overwrite & $36.0\,\pm\,10.8$ & $50.1\,\pm\,8.7$ & $69.0\,\pm\,4.2$ \\
A5V5 & GDN & retention & $64.9\,\pm\,32.1$ & --- & $88.0\,\pm\,8.9$ \\
A5V5 & GDN & interference & $60.8\,\pm\,32.0$ & $66.3\,\pm\,32.1$ & $86.7\,\pm\,9.9$ \\
A5V5 & GDN & overwrite & $59.4\,\pm\,31.0$ & $65.8\,\pm\,26.6$ & $84.7\,\pm\,11.5$ \\
\midrule
\end{longtable}\endgroup

\clearpage
\section{Dynamic-State Ablation}
\label{app:state-ablation}

\paragraph{Intervention and controls.}
We evaluate all 27 final \sm{} checkpoints from the main comparison
(A3--A5, V3--V5, three training seeds) with and without their dynamic states.
In the ablated condition, every layer's $S$ is set to zero immediately after
\emph{every} periodic or write/read-segment-boundary commit, so subsequent reads
always use $E+0$. This is not a one-time reset followed by normal accumulation.
Weights, static schema embeddings $E$, write computations, chunk-local attention,
position counters, and the commit schedule are unchanged; no retraining is used.

We pair the two conditions on identical episodes at delays 384 and 1,536,
using 512 episodes per scenario, batch size 128, and evaluation seed 20261001.
Only final queries are scored, as in the main evaluation.
Recomputed normal-condition scores match the frozen main results exactly.
Tables~\ref{tab:state-ablation-384} and~\ref{tab:state-ablation-1536} report
exact-value accuracy as mean $\pm$ sample SD over the three training seeds.

\paragraph{Results and interpretation.}
Retained- and updated-value accuracy falls below 1\% for every checkpoint in
both delay conditions. For example, A3V3 overwritten-value accuracy falls from
95.35\% to 0\% at delay 384 and from 74.64\% to 0\% at delay 1,536.
Default predictions are largely preserved: across all 27 checkpoints, their
mean accuracy in the overwrite scenario is 99.52\% with $S=0$ versus 99.83\%
normally at delay 384. At delay 1,536 it rises from 64.15\% to 99.51\%, with
an improvement in every checkpoint.
Written values are sampled to differ from their defaults, so reverting to the
learned default mapping does not recover the episode-specific answers.
The intervention therefore verifies that the evaluated models use $S$ to convey
written values across chunks, while also exposing state-dependent errors on
untouched addresses. It does not isolate the benefit of learned schema embeddings
from other memory parameterizations.

\begin{table}[htp]\centering
\caption{SchemaMem dynamic-state ablation at delay 384: exact-value accuracy (\%, mean $\pm$ sample SD over three training seeds). A$a$V$v$ denotes $a$-bit addresses ($2^a$ addresses) and $v$-bit values ($2^v$ possible values). Normal and $S=0$ use the same checkpoints and episodes. $S=0$ clears all layers' dynamic states after every commit. Retained-R: retention-only; Retained-I: retained after disjoint writes; Updated and Default: overwrite scenario.}
\label{tab:state-ablation-384}
\begin{tabular}{llrrrr}\toprule
Setting & State & Retained-R & Retained-I & Updated & Default \\\midrule
A3V3 & Normal & $99.1\,\pm\,1.4$ & $99.1\,\pm\,1.1$ & $95.3\,\pm\,0.3$ & $99.9\,\pm\,0.1$ \\
A3V3 & $S=0$ & $0.0\,\pm\,0.0$ & $0.0\,\pm\,0.0$ & $0.0\,\pm\,0.0$ & $100.0\,\pm\,0.0$ \\
\midrule
A3V4 & Normal & $99.4\,\pm\,0.7$ & $99.1\,\pm\,1.5$ & $96.1\,\pm\,1.3$ & $99.9\,\pm\,0.0$ \\
A3V4 & $S=0$ & $0.0\,\pm\,0.0$ & $0.0\,\pm\,0.0$ & $0.0\,\pm\,0.0$ & $100.0\,\pm\,0.0$ \\
\midrule
A3V5 & Normal & $99.9\,\pm\,0.2$ & $99.9\,\pm\,0.2$ & $96.4\,\pm\,1.0$ & $100.0\,\pm\,0.1$ \\
A3V5 & $S=0$ & $0.0\,\pm\,0.0$ & $0.0\,\pm\,0.0$ & $0.0\,\pm\,0.0$ & $100.0\,\pm\,0.0$ \\
\midrule
A4V3 & Normal & $99.7\,\pm\,0.5$ & $99.8\,\pm\,0.4$ & $95.4\,\pm\,0.3$ & $99.9\,\pm\,0.3$ \\
A4V3 & $S=0$ & $0.0\,\pm\,0.0$ & $0.0\,\pm\,0.0$ & $0.0\,\pm\,0.0$ & $100.0\,\pm\,0.0$ \\
\midrule
A4V4 & Normal & $99.9\,\pm\,0.1$ & $99.9\,\pm\,0.1$ & $96.0\,\pm\,1.6$ & $99.8\,\pm\,0.1$ \\
A4V4 & $S=0$ & $0.0\,\pm\,0.0$ & $0.0\,\pm\,0.0$ & $0.0\,\pm\,0.0$ & $100.0\,\pm\,0.0$ \\
\midrule
A4V5 & Normal & $99.7\,\pm\,0.3$ & $99.9\,\pm\,0.2$ & $95.3\,\pm\,0.4$ & $99.9\,\pm\,0.2$ \\
A4V5 & $S=0$ & $0.1\,\pm\,0.2$ & $0.1\,\pm\,0.2$ & $0.2\,\pm\,0.3$ & $96.1\,\pm\,6.7$ \\
\midrule
A5V3 & Normal & $99.7\,\pm\,0.2$ & $99.8\,\pm\,0.1$ & $95.6\,\pm\,0.6$ & $99.6\,\pm\,0.3$ \\
A5V3 & $S=0$ & $0.0\,\pm\,0.0$ & $0.0\,\pm\,0.0$ & $0.0\,\pm\,0.0$ & $100.0\,\pm\,0.0$ \\
\midrule
A5V4 & Normal & $99.8\,\pm\,0.1$ & $99.7\,\pm\,0.1$ & $95.7\,\pm\,0.2$ & $99.8\,\pm\,0.0$ \\
A5V4 & $S=0$ & $0.0\,\pm\,0.1$ & $0.1\,\pm\,0.1$ & $0.0\,\pm\,0.0$ & $99.6\,\pm\,0.7$ \\
\midrule
A5V5 & Normal & $99.8\,\pm\,0.2$ & $99.9\,\pm\,0.1$ & $96.8\,\pm\,0.3$ & $99.6\,\pm\,0.3$ \\
A5V5 & $S=0$ & $0.0\,\pm\,0.0$ & $0.0\,\pm\,0.0$ & $0.0\,\pm\,0.0$ & $100.0\,\pm\,0.0$ \\
\bottomrule\end{tabular}\end{table}

\clearpage
\begin{table}[htp]\centering
\caption{SchemaMem dynamic-state ablation at delay 1,536: exact-value accuracy (\%, mean $\pm$ sample SD over three training seeds). A$a$V$v$ denotes $a$-bit addresses ($2^a$ addresses) and $v$-bit values ($2^v$ possible values). Normal and $S=0$ use the same checkpoints and episodes. $S=0$ clears all layers' dynamic states after every commit. Retained-R: retention-only; Retained-I: retained after disjoint writes; Updated and Default: overwrite scenario.}
\label{tab:state-ablation-1536}
\begin{tabular}{llrrrr}\toprule
Setting & State & Retained-R & Retained-I & Updated & Default \\\midrule
A3V3 & Normal & $81.4\,\pm\,5.3$ & $78.9\,\pm\,3.6$ & $74.6\,\pm\,5.8$ & $71.3\,\pm\,10.6$ \\
A3V3 & $S=0$ & $0.0\,\pm\,0.0$ & $0.0\,\pm\,0.0$ & $0.0\,\pm\,0.0$ & $100.0\,\pm\,0.0$ \\
\midrule
A3V4 & Normal & $71.9\,\pm\,9.3$ & $71.2\,\pm\,9.1$ & $66.3\,\pm\,8.8$ & $71.7\,\pm\,13.5$ \\
A3V4 & $S=0$ & $0.0\,\pm\,0.0$ & $0.0\,\pm\,0.0$ & $0.0\,\pm\,0.0$ & $100.0\,\pm\,0.0$ \\
\midrule
A3V5 & Normal & $72.5\,\pm\,25.3$ & $70.2\,\pm\,27.5$ & $65.8\,\pm\,21.8$ & $64.8\,\pm\,21.7$ \\
A3V5 & $S=0$ & $0.0\,\pm\,0.0$ & $0.0\,\pm\,0.0$ & $0.0\,\pm\,0.0$ & $100.0\,\pm\,0.0$ \\
\midrule
A4V3 & Normal & $80.2\,\pm\,24.2$ & $78.1\,\pm\,24.8$ & $77.7\,\pm\,18.3$ & $62.8\,\pm\,14.3$ \\
A4V3 & $S=0$ & $0.0\,\pm\,0.0$ & $0.0\,\pm\,0.0$ & $0.0\,\pm\,0.0$ & $100.0\,\pm\,0.0$ \\
\midrule
A4V4 & Normal & $82.0\,\pm\,6.4$ & $79.3\,\pm\,5.9$ & $75.6\,\pm\,1.8$ & $62.1\,\pm\,15.3$ \\
A4V4 & $S=0$ & $0.0\,\pm\,0.0$ & $0.0\,\pm\,0.0$ & $0.0\,\pm\,0.0$ & $100.0\,\pm\,0.0$ \\
\midrule
A4V5 & Normal & $76.9\,\pm\,12.5$ & $76.0\,\pm\,16.9$ & $74.0\,\pm\,15.7$ & $68.9\,\pm\,3.3$ \\
A4V5 & $S=0$ & $0.1\,\pm\,0.2$ & $0.1\,\pm\,0.1$ & $0.2\,\pm\,0.4$ & $96.1\,\pm\,6.7$ \\
\midrule
A5V3 & Normal & $69.5\,\pm\,8.9$ & $66.1\,\pm\,9.4$ & $64.3\,\pm\,12.4$ & $59.4\,\pm\,2.1$ \\
A5V3 & $S=0$ & $0.0\,\pm\,0.0$ & $0.0\,\pm\,0.0$ & $0.0\,\pm\,0.0$ & $100.0\,\pm\,0.0$ \\
\midrule
A5V4 & Normal & $68.7\,\pm\,11.8$ & $65.7\,\pm\,12.0$ & $64.5\,\pm\,10.3$ & $59.5\,\pm\,2.5$ \\
A5V4 & $S=0$ & $0.0\,\pm\,0.0$ & $0.0\,\pm\,0.1$ & $0.0\,\pm\,0.1$ & $99.5\,\pm\,0.8$ \\
\midrule
A5V5 & Normal & $72.3\,\pm\,10.6$ & $72.5\,\pm\,7.6$ & $67.4\,\pm\,4.3$ & $56.9\,\pm\,8.3$ \\
A5V5 & $S=0$ & $0.0\,\pm\,0.0$ & $0.0\,\pm\,0.0$ & $0.0\,\pm\,0.0$ & $100.0\,\pm\,0.0$ \\
\bottomrule\end{tabular}\end{table}

\end{document}